\documentclass[11pt]{article}

\usepackage[final]{acl}

\usepackage{times}
\usepackage{latexsym}

\usepackage[T1]{fontenc}

\usepackage[utf8]{inputenc}

\usepackage{microtype}

\usepackage{inconsolata}

\usepackage{graphicx}

\usepackage{multirow}
\usepackage{tcolorbox}
\usepackage{placeins}
\usepackage{booktabs}
\usepackage{enumitem}
\usepackage{tikz}
\usepackage{enumitem}
\usepackage{fontawesome5}
\usepackage{amsmath}
\usepackage{amsfonts}
\usetikzlibrary{positioning, shapes, arrows.meta, shadows.blur}

\title{High Scores, Hidden Failures:\\ Evaluating Codebook-Based LLM Content Analysis}

\author{First Author \\
  Affiliation / Address line 1 \\
  Affiliation / Address line 2 \\
  Affiliation / Address line 3 \\
  \texttt{email@domain} \\\And
  Second Author \\
  Affiliation / Address line 1 \\
  Affiliation / Address line 2 \\
  Affiliation / Address line 3 \\
  \texttt{email@domain} \\}

\author{Elena Senger$^{1,2}$ \quad Yuri Campbell$^{2}$ \quad Rob van der Goot$^{3}$ \quad Barbara Plank$^{1}$ \\[1.5ex]
$^1$MaiNLP, Center for Information and Language Processing, LMU Munich \& MCML, Germany \\
$^2$Fraunhofer Institute for Systems and Innovation Research ISI, Germany\\
$^3$Department of Computer Science, IT University of Copenhagen, Denmark \\
\texttt{elena.senger@cis.lmu.de, yuri.campbell@isi.fraunhofer.de} \\
\texttt{robv@itu.dk, b.plank@lmu.de}
}

\begin{document}
\maketitle
\begin{abstract}

Large language models (LLMs) are increasingly used for codebook-based content analysis in computational social science, where codebooks translate theoretical concepts into labeling schemes. While prior work reports strong predictive performance, it remains unclear 
whether these systems produce reliable and robust text annotations. 
We conduct a large-scale empirical failure-mode analysis of LLM-based coding pipelines across four datasets and five models by systematically evaluating their stability, robustness, and conceptual validity under controlled perturbations, domain shifts, and optimization procedures. 
We uncover multiple failure modes that are largely invisible under standard metrics, showing
that high aggregate scores frequently coexist with severe label-level disparities, instability across runs, domain sensitivity, and shifts in label interpretation.
Importantly, these failure modes persist under widely used optimization strategies and remain largely invisible under standard metrics such as accuracy and micro-F1. 
Our results demonstrate that current LLM-based coding pipelines cannot be treated as stable measurement tools without targeted validation beyond aggregate performance. 
\end{abstract}

\section{Introduction}
Large language models (LLMs) are increasingly used to annotate text data using natural language instructions, enabling scalable labeling for tasks such as dataset construction, qualitative coding, and content analysis~\citep{Toernberg_2024, abraham2025promptselectionmattersenhancing, rytting2023codingsocialsciencedatasets, xiao_23}. In computational social science (CSS), these instructions are typically formalized as codebooks, structured annotation guidelines that map theoretical concepts onto predefined labels. For example, a codebook for analysing news coverage of minimum wage policies might define labels such as \textit{Prices}, \textit{Government}, \textit{Welfare}, and \textit{Justice} to classify the different concepts through which minimum wage policies are discussed in each text. By incorporating 
 the codebook's labels and their definitions
directly into prompts, recent studies report strong predictive performance and substantial agreement between human and LLM annotations, suggesting that LLMs could serve as scalable substitutes for human coders ~\citep{rytting2023codingsocialsciencedatasets, Ramanathan_2025, ruckdeschel-2025-just, farjam_2025}.

\begin{figure}
  \centering
  \includegraphics[width=0.8\linewidth]{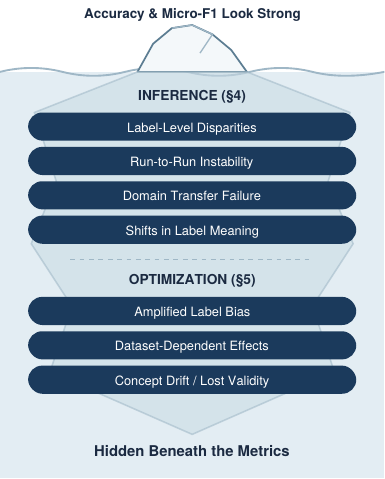}
  \caption{Illustrative examples of LLM failure modes in inference and optimization hidden beneath metrics.}
  \label{fig:fig1}
\end{figure}

However, the use of LLMs for codebook-based annotation introduces two sources of potential failure. First, LLMs may fail to apply codebook instructions consistently during inference. A growing body of NLP research highlights prompt sensitivity~\citep{shu-etal-2024-dont, Mizrahi_2024, gu-etal-2023-robustness}, where model predictions change under seemingly minor prompt variations, such as formatting choices, label ordering, or wording differences. Beyond prompt sensitivity, 
LLM-based coding pipelines may also exhibit broader robustness issues, including instability across repeated runs, inconsistent adherence to codebook definitions, and large disparities in performance across individual labels.
Second, LLMs are increasingly used not only to apply annotation instructions but also to refine or optimize the prompts themselves. Recent approaches employ automatic prompt optimization or multi-agent prompting frameworks to iteratively modify prompts in order to improve predictive performance~\citep{abraham2025promptselectionmattersenhancing, zhao-etal-2025-scale}. While these methods can yield improvements on standard evaluation metrics, they also introduce additional risks. For instance, modifying prompts that encode coding schemes can implicitly alter label definitions, amplify disparities in label-level performance, or shift the conceptual meaning of the constructs being measured (see Section~\ref{sec:optimization}).
These issues are consequential in content analysis, where annotations function as measurements ~\citep{krippendorff2018content}. They feed downstream statistical analyses (frequency comparisons, regressions, hypothesis testing), as well as qualitative interpretation, including conceptual and relational analysis. This requires that performance is balanced across labels for frequency-based claims to hold~\citep{egami2023using}, results are comparable across runs, models, and corpora for findings to replicate, and the labels must retain the theoretical constructs the codebook was designed to measure~\citep{halterman2025codebookllmsevaluatingllms}. 

In this paper, we operationalize the risks around faithful LLM coding by defining controlled experiments and complementary metrics. Then, we empirically demonstrate that current LLM-based codebook methods systematically fail in predictable ways.
We organize our study around a set of \emph{failure modes}, spanning two stages of practical CSS pipelines: (1) \textbf{inference failure modes}, which arise when an LLM applies a fixed codebook (see Section~\ref{sec:inference}), and (2) \textbf{optimization-induced failure modes}, which arise when an LLM is used to refine a codebook prior to inference (see Section~\ref{sec:optimization}).
Our main contributions are:
\begin{itemize}[topsep=0.5em, itemsep=0.3em, parsep=0em, partopsep=0em]
\item A framework for systematic failure-mode analysis for LLM content analysis, identifying predictable breakdowns at both the inference and optimization stages.
\item Evidence across four datasets and five models that commonly reported performance metrics can misrepresent coding reliability by masking these failures, along with a minimal validation protocol (see Section~\ref{sec:protocol}) to uncover such failure modes.
\item Two new expert-annotated datasets sharing a single codebook, enabling controlled domain-transfer evaluation, as well as code and evaluation pipeline to support reproducibility.\footnote{\textsc{Value creation Science} dataset:~\url{https://huggingface.co/datasets/ElenaSenger/value-creation-science}; \textsc{Value creation News} dataset:~\url{https://huggingface.co/datasets/ElenaSenger/value-creation-news}; Code:~\url{https://github.com/elenasenger/high-scores-hidden-failures}}
\end{itemize}

\section{Related Work}

\subsection{Reliability of LLM-Based Text Annotation}
In CSS, LLMs are increasingly used to generate annotations for downstream analyses, such as estimating prevalence or testing hypotheses \citep{gligoric-etal-2025-unconfident, baumann2025largelanguagemodelhacking}. It has been shown that LLM annotations vary depending on the design of the prompt, the domain, the task, or the choice of model \citep{aldeen_2023, KOCON2023}. Therefore, naive use can produce unreliable statistical estimates \citep{gligoric-etal-2025-unconfident} and propagate to downstream statistical inference \citep{baumann2025largelanguagemodelhacking}. 

Within content analysis, research has increasingly explored how LLMs can leverage codebooks in zero-shot or few-shot coding scenarios \citep{rytting2023codingsocialsciencedatasets, dunivin2025scaling, farjam_2025}. Prior work shows that incorporating codebook definitions or examples can improve classification performance \citep{ruckdeschel-2025-just, xiao_23}. Other studies highlight that prompt design choices and codebook components (e.g., “Negative Clarification” or “Positive Example”) can influence model predictions \citep{halterman2025codebookllmsevaluatingllms}. Some work also reports variability in LLM coding results across repeated runs despite strong overall agreement with human annotations \citep{Ramanathan_2025}, suggesting that stability remains an underexplored dimension of evaluation. However, existing studies primarily focus on aggregate performance metrics \citep{ruckdeschel-2025-just, xiao_23, rytting2023codingsocialsciencedatasets, dunivin2025scaling, farjam_2025}, leaving open questions about robustness, label-level disparities, and other potential failure modes when LLMs apply codebooks.

Where prior work documents that predictions shift under prompt variation, we systematically evaluate published codebook-design recommendations  \citep{ruckdeschel-2025-just, halterman2025codebookllmsevaluatingllms, dunivin2025scaling}, run-to-run stability, label-name reliance, and domain transfer under controlled conditions. Our analysis extends prior findings by decomposing these effects at the label level and distinguishing robustness to stylistic variation from adherence to the underlying coding concepts.


\subsection{Codebook Optimization with LLMs}

A complementary line of research explores how LLMs themselves can be used to optimize codebook-prompts for LLM inference. 
\citet{abraham2025promptselectionmattersenhancing} propose Automatic Prompt Optimization (APO), starting with a  human-labeled subset and an initial task prompt. The LLM repeatedly generates semantically equivalent variations, evaluates them on the labeled data, and retains the best-performing prompts. 
Additionally, \citet{zhao-etal-2025-scale} introduce the multi-agent framework SCALE. SCALE simulates collaborative coding processes, including discussion and evolving codebooks. 
In our study, we apply APO and SCALE, analysing their robustness, comparability, and impact on the quality of codebooks.
While prompt optimization techniques focus on improving predictive performance, their effects on robustness, and conceptual validity remain unexplored. In this work, we systematically analyse how codebook optimization interact with LLM-based annotation pipelines and show that performance improvements can coincide with amplified failure modes.

\section{Data and Evaluation Setup}
\label{sec:datasets_setups}

\paragraph{Datasets}
To address robustness, we evaluate across four datasets varying in domain, size, and label count (Table~\ref{tab:datasets}; codebooks in Appendix~\ref{app:codebooks}). We include the \textsc{CCC} dataset \citep{halterman2025codebookllmsevaluatingllms} (on protests in the United States) and  \textsc{Minimum Wage} dataset \cite{mattes_2023_dataset}  (on the impact of minimum wage policies) from prior work to ensure comparability. 
In addition, we introduce and release two datasets on Value Creation across distinct sources: the \textsc{Value Creation Science} dataset, derived from scientific literature, and the \textsc{Value Creation News}, compiled from news-wire corpora.
The former is the primary dataset and the latter was developed to evaluate robustness under domain shift.
The related codebook 
captures concepts such as actor, demand, and resource, and was developed by a team of social scientists through an iterative annotation process, with refinements guided by inter-annotator agreement (Fleiss' $\kappa$). Two domain experts further provide qualitative validation for this present study. 
Additional details on data sources, label sets, and annotation procedures are provided in Appendix~\ref{app:datasets}.

\begin{table}
\centering
\small
\resizebox{\columnwidth}{!}{%
\begin{tabular}{lcrrrr}

\hline
\textbf{Dataset} & \textbf{Multi-Label}  & \textbf{\#Labels} & \textbf{\#Train} & \textbf{\#Val} & \textbf{\#Test} \\ 
\hline
\textsc{CCC} & $\times$ & 4 & 4,706 & 1,007 & 1,010 \\
\textsc{Minimum Wage} & $\checkmark$ & 12  & 758 & 194 & 194 \\
\textsc{Value Creation Science} & $\checkmark$ & 8 & 2,149 & 268 & 269 \\
\textsc{Value Creation News} & $\checkmark$ & 8 & -- & -- & 60 \\
\hline
\end{tabular}
}
\caption{Dataset statistics. Data split size is in number of documents.}
\label{tab:datasets}
\end{table}

\renewcommand{\arraystretch}{1.2}
\begin{table*}
\centering
\resizebox{\textwidth}{!}{%
\addtolength{\tabcolsep}{-0.1cm}

\begin{tabular}{l|ccccc|ccccc|ccccc}
\hline
\multirow{2}{*}{\textbf{Codebook Variant}} 
 & \multicolumn{5}{c}{\textbf{\textsc{Minimum Wage} }} 
 & \multicolumn{5}{|c|}{\textbf{\textsc{CCC}}} 
 & \multicolumn{5}{c}{\textbf{\textsc{Value Creation Science}}} \\
\cline{2-16}
 & \textbf{Acc} & \textbf{mi.\ F1} & \textbf{ma.\ F1} & \textbf{ma.\ P} & \textbf{WLF1}
 & \textbf{Acc} & \textbf{mi.\ F1} & \textbf{ma.\ F1} & \textbf{ma.\ P} & \textbf{WLF1}
 & \textbf{Acc} & \textbf{mi.\ F1} & \textbf{ma.\ F1} & \textbf{ma.\ P} & \textbf{WLF1}
 \\
\hline
Unchanged (§\ref{sec:label_level}, §\ref{sec:overinterpretation})    
 & $18.2_{16.8}^{19.9}$ & $59.1_{58.2}^{59.9}$ & $54.8_{53.2}^{56.4}$ & $47.5_{45.8}^{49.2}$ & $28.9_{19.1}^{38.7}$ 
 & $33.0_{32.7}^{33.2}$ & $34.0_{33.8}^{34.3}$ & $33.4_{33.1}^{33.7}$ & $45.6_{45.4}^{45.8}$ & $6.4_{6.4}^{6.5}$
 & $1.9_{1.6}^{2.1}$ & $44.9_{44.6}^{45.1}$ & $39.6_{39.0}^{40.1}$ & $30.5_{29.9}^{31.0}$ & $18.8_{15.9}^{21.7}$
 \\ 
\hline
Generic Label (§\ref{sec:attention})             
 & $\underline{\mathbf{23.4_{22.0}^{24.9}}}$ & $\underline{\mathbf{55.7_{55.1}^{56.2}}}$ & $\underline{\mathbf{49.5_{48.5}^{50.6}}}$ & $\underline{\mathbf{39.7_{38.4}^{41.0}}}$ & $32.2_{24.5}^{39.9}$
 & $\underline{\mathbf{42.9_{42.6}^{43.1}}}$ & $\underline{\mathbf{43.8_{43.5}^{44.1}}}$ & $\underline{\mathbf{27.8_{27.6}^{28.0}}}$ & $45.8_{45.1}^{46.5}$ & $\underline{\mathbf{5.8_{5.7}^{6.0}}}$
 & $\mathbf{1.0_{0.7}^{1.2}}$ & $\underline{\mathbf{41.6_{41.2}^{42.1}}}$ & $\mathbf{38.7_{38.1}^{39.2}}$ & $\mathbf{27.5_{27.1}^{27.9}}$ & $19.3_{17.3}^{21.4}$
 \\ 
Swapped Label (§\ref{sec:attention})   
 & $\underline{\mathbf{2.9_{2.4}^{3.4}}}$ & $\underline{\mathbf{37.7_{37.0}^{38.5}}}$ & $\underline{\mathbf{29.9_{28.6}^{31.2}}}$ & $\underline{\mathbf{25.2_{24.0}^{26.4}}}$ & $\underline{\mathbf{1.4_{0.0}^{4.7}}}$
 & $\underline{\mathbf{12.3_{12.2}^{12.4}}}$ & $\underline{\mathbf{12.5_{12.4}^{12.7}}}$ & $\underline{\mathbf{10.6_{10.3}^{10.8}}}$ & $\underline{\mathbf{17.4_{17.1}^{17.6}}}$ & $\underline{\mathbf{2.5_{2.3}^{2.6}}}$
 & $2.0_{1.8}^{2.2}$ & $\underline{\mathbf{39.9_{39.4}^{40.4}}}$ & $\underline{\mathbf{35.1_{34.6}^{35.7}}}$ & $\mathbf{28.5_{28.0}^{28.9}}$ & $20.6_{19.6}^{21.6}$
 \\ 
Label-Only (§\ref{sec:attention})   
 & $\underline{\mathbf{6.9_{6.4}^{7.3}}}$ & $\underline{\mathbf{45.7_{45.5}^{45.9}}}$ & $\underline{\mathbf{42.1_{41.8}^{42.4}}}$ & $\underline{\mathbf{36.2_{35.9}^{36.5}}}$ & $\underline{\mathbf{14.2_{13.3}^{15.1}}}$
 & $\underline{\mathbf{15.9_{15.6}^{16.1}}}$ & $\underline{\mathbf{16.8_{16.6}^{17.0}}}$ & $\underline{\mathbf{24.0_{23.7}^{24.7}}}$ & $\mathbf{46.5_{46.0}^{46.9}}$ & $\underline{\mathbf{5.8_{5.8}^{5.8}}}$
 & $\mathbf{0.6_{4.5}^{8.1}}$ & $\mathbf{43.5_{43.2}^{43.8}}$ & $\mathbf{38.4_{37.9}^{38.8}}$ & $\mathbf{29.2_{28.8}^{29.6}}$ & $\underline{\mathbf{24.2_{22.0}^{26.4}}}$
 \\ [2pt]
\hline
Simple Codebook (§\ref{sec:guidelines}.i)     
 & $\mathbf{21.6_{21.1}^{22.1}}$ & $\mathbf{61.2_{60.7}^{61.7}}$ & $\mathbf{56.8_{55.9}^{57.8}}$ & $48.9_{47.9}^{49.9}$ & $\underline{\mathbf{39.1_{38.0}^{40.1}}}$
 & $\underline{\mathbf{30.0_{26.7}^{27.2}}}$ & $\underline{\mathbf{28.7_{28.4}^{28.9}}}$ & $\underline{\mathbf{31.2_{31.0}^{31.4}}}$ & $45.6_{45.5}^{45.8}$ & $\underline{\mathbf{6.1_{5.9}^{6.3}}}$
 & $1.7_{1.5}^{2.0}$ & $\underline{\mathbf{42.5_{42.3}^{42.8}}}$ & $\underline{\mathbf{35.9_{35.5}^{36.3}}}$ & $\underline{\mathbf{26.7_{26.2}^{27.2}}}$ & $\underline{\mathbf{10.9_{8.6}^{13.3}}}$
 \\ 
Structured (§\ref{sec:guidelines}.ii)            
 & $\mathbf{21.0_{19.2}^{22.8}}$ & $\underline{\mathbf{56.8_{55.8}^{57.7}}}$ & $\underline{\mathbf{50.9_{49.5}^{52.4}}}$ & $\underline{\mathbf{43.3_{41.7}^{45.0}}}$ & $25.8_{14.9}^{36.7}$
 & $\underline{\mathbf{49.3_{49.1}^{49.5}}}$ & $\underline{\mathbf{50.2_{50.0}^{50.3}}}$ & $\underline{\mathbf{38.6_{38.5}^{38.8}}}$ & $\mathbf{47.4_{47.2}^{47.6}}$ & $\underline{\mathbf{7.0_{6.8}^{7.2}}}$
 & $1.9_{1.5}^{2.2}$ & $\mathbf{44.3_{44.0}^{44.7}}$ & $\mathbf{38.1_{37.7}^{38.6}}$ & $29.8_{29.0}^{30.6}$ & $\underline{\mathbf{12.1_{10.0}^{14.1}}}$
 \\ 
Majority Order (§\ref{sec:guidelines}.iii)    
 & $\mathbf{20.7_{19.5}^{21.9}}$ & $60.2_{59.2}^{61.1}$ & $54.7_{53.1}^{56.2}$ & $45.7_{44.0}^{47.4}$ & $36.4_{31.8}^{41.0}$
 & $\underline{\mathbf{29.0_{28.9}^{29.2}}}$ & $\underline{\mathbf{30.1_{30.0}^{30.3}}}$ & $\mathbf{32.0_{31.8}^{32.1}}$ & $\mathbf{46.7_{46.4}^{47.0}}$ & $\underline{\mathbf{6.2_{6.1}^{6.4}}}$
 & $\underline{\mathbf{0.4_{0.3}^{0.6}}}$ & $\mathbf{43.6_{43.3}^{43.9}}$ & $40.2_{39.8}^{40.6}$ & $29.9_{29.6}^{30.2}$ & $21.8_{21.0}^{22.5}$
 \\ 
Inverted Order (§\ref{sec:guidelines}.iii)    
 & $\underline{\mathbf{13.1_{12.2}^{14.0}}}$ & $\underline{\mathbf{55.9_{55.4}^{56.4}}}$ & $\underline{\mathbf{49.8_{49.4}^{50.3}}}$ & $\underline{\mathbf{42.3_{41.8}^{42.7}}}$ & $31.8_{30.9}^{32.7}$
 & $\underline{\mathbf{57.7_{57.5}^{57.9}}}$ & $\underline{\mathbf{58.7_{58.6}^{58.9}}}$ & $\underline{\mathbf{45.4_{45.2}^{45.5}}}$ & $\mathbf{47.0_{46.8}^{47.1}}$ & $\underline{\mathbf{9.0_{8.5}^{9.5}}}$
 & $\mathbf{0.7_{0.4}^{1.0}}$ & $\underline{\mathbf{36.2_{35.8}^{36.6}}}$ & $\underline{\mathbf{33.3_{32.7}^{33.8}}}$ & $\mathbf{33.4_{32.5}^{34.3}}$ & $16.3_{14.9}^{17.6}$
\\[2pt]
\hline
\end{tabular}%
}
\caption{LLM-Inference results with baseline Unchanged Codebook vs. other codebook variants. We report accuracy (Acc), micro-F1 (mi.\ F1), macro-F1 (ma.\ F1), macro precision (ma.\ P), and worst-label-F1 (WLF1). 
For all metrics, we show their mean value and the 95\% confidence interval over 10 runs.
Bold values indicate statistically significant differences over baseline ($p<0.05$) under paired $t$-tests. 
Underlined values indicate significance under paired bootstrap tests.
Codebook variant numbers (e.g., §\ref{sec:attention}) indicate where each variant is discussed. Results of an encoder baseline can be found in Appendix~\ref{app:roberta_baseline} for comparison. }
\label{tab:label_attendance}
\end{table*}

\paragraph{Models and Evaluation} 
Unless otherwise noted, inference experiments use Mistral-7B-Instruct-v0.3 \citep{jiang2023mistral7b} at temperature 0.3, an open and relatively compact model accessible to a wide range of researchers.\footnote{
We use a temperature of 0.3 to enable the run-to-run variability observed in prior work and to stay close to the provider defaults and chat interfaces that CSS practitioners typically rely on~\citep{tai2024examination}.}
Cross-model comparisons appear in Section~\ref{sec:model_gen}. To ensure comparability with the original SCALE setup, we run all three methods with GPT-4o. 
We report \textit{accuracy}, \textit{micro-F1}, \textit{macro-F1}, \textit{macro-Precision}, and \textit{Worst-Label-F1} to surface disparities masked by aggregate metrics. Accuracy measures the fraction of instances where every label of that instance is predicted correctly. For codebook-optimization approaches, we additionally report a human expert evaluation with Likert scale  to evaluate the resulting codebook quality on \textsc{Value Creation Science}.

\section{Failure Modes in Inference}
\label{sec:inference}

In the following subsections, we present the results of the inference experiments structured along their respective identified failure modes. Starting from an \textit{Unchanged} codebook baseline (Table~\ref{tab:label_attendance}), we examine issues such as label-level disparities and overinterpretation. We further probe reliability through prompt diagnostics, evaluate context sensitivity via formatting and dataset variations, measure stochastic stability with repeated runs, test domain transfer on a new corpus, and assess model generalization across five LLMs.

\subsection{Reliability of Label Application}

\paragraph{Label-Level Disparities}
\label{sec:label_level}

Standard metrics such as \textit{accuracy} and \textit{micro-F1} aggregate across labels and can therefore mask severe per-category failures, a problem compounded by the highly imbalanced datasets typical of CSS. Balanced performance is essential  for valid downstream interpretation. For example, in a simple frequency analysis of how often a concept appears in a corpus, systematic underperformance on certain labels directly biases results. Moreover, robust generalization to datasets with different label distributions requires models to perform reliably across all categories. 
Our experiments expose this pattern (Table~\ref{tab:label_attendance}). In \textsc{Minimum Wage}, the label \textsc{Youth-and-Secondary-Wage-Earners} (support\,=\,5) achieves the lowest F1 ($28.90$), while \textsc{Social-Justice/Injustice} (support\,=\,59) reaches 72.97. More strikingly, the gap between \textit{worst-label-F1} and \textit{micro-F1} is 30 points for \textsc{Minimum Wage}, 26 points for \textsc{Value Creation Science}, and 28 points for \textsc{CCC}, which are differences entirely invisible in aggregate reporting. 
The consistent gap between \textit{macro-F1} and \textit{micro-F1} across all experiments further confirms uneven performance.

\paragraph{Overinterpretation}
\label{sec:overinterpretation}
\citet{ruckdeschel-2025-just} identify overinterpretation as a recurring issue in LLM-based coding: models infer meaning beyond what is explicitly stated. Our quantitative results confirm this tendency. In two of the three datasets, \textit{macro-Precision} falls below \textit{macro-F1} (Table~\ref{tab:label_attendance}), indicating that models assign labels more liberally than the gold standard warrants. 
At the label level, precision lags behind recall in 17 of 24 labels, pointing to systematic overprediction rather than isolated errors.
For CSS workflows that rely on label frequencies or co-occurrence counts \citep{baumann2025largelanguagemodelhacking, gligoric-etal-2025-unconfident}, such overcoding inflates the apparent prevalence of concepts in a corpus and can distort substantive conclusions.

\paragraph{Codebook-Specific Label Meanings} 
\label{sec:attention}

A central assumption of codebook-based content analysis is that the model follows the provided definitions rather than defaulting to background knowledge acquired during pretraining. \citet{ziems-etal-2024-large} argue that zero-shot performance is strongest when a model’s internal concept overlaps with the intended construct (e.g., stance or emotion detection), but declines on tasks requiring compliance with detailed instructions. \citet{halterman2025codebookllmsevaluatingllms} make this issue explicit with two diagnostics: \textit{Generic Label} Accuracy (replacing labels with pseudo-labels such as ``Label 1'') and \textit{Swapped Label} Accuracy (permuting names and definitions). 
Building on this work, we introduce a third diagnostic, \textit{Label-Only}, in which inference uses only label names, without their definitions. 
The results (Table~\ref{tab:label_attendance}) reveal a consistent pattern across datasets: \textit{Label-Only} achieves surprisingly high scores (Table~\ref{tab:label_attendance}), especially for \textsc{Value Creation Science}, suggesting that models lean heavily on label names.  In contrast, performance drops substantially under \textit{Generic Label} and \textit{Swapped Label}, sometimes more than in the \textit{Label-Only} case, confirming reliance on label name over definitions.

\subsection{Robustness}
\label{sec:robustness}

\begin{figure*}
  \centering
  \includegraphics[width=1.0\linewidth]{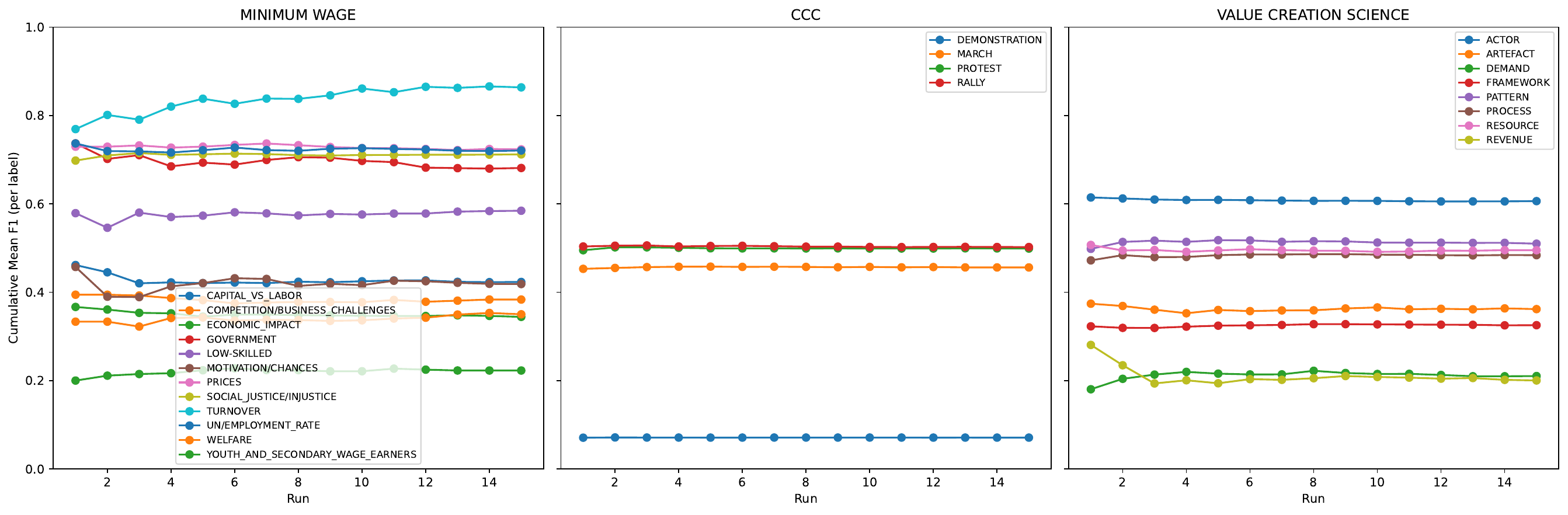}
  \caption{Aggregated mean F$_1$ scores over 15 runs, illustrating the stabilizing effect of averaging multiple runs for each label.}
  \label{fig:rerun_label}
\end{figure*}
\paragraph{Context Sensitivity}
\label{sec:guidelines}

Prior work offers various, often conflicting guidelines for prompt design in LLM-based content analysis. We isolate three such recommendations, applying each as a single modification to the \textit{Unchanged} codebook baseline (Table~\ref{tab:label_attendance}).

\citet{ruckdeschel-2025-just} advocates for shorter, concise codebooks, while \citet{halterman2025codebookllmsevaluatingllms} suggest using full codebooks for better overall performance.
In our experiments, we use labels and their definitions as simplified codebooks, removing examples, clarifications and counter-examples (\textit{i. Simple} codebook, Table~\ref{tab:label_attendance}). 
We see that simplified codebooks degrade both \textit{micro/macro-F1} for \textsc{Value Creation Science} and \textsc{CCC} datasets, but improve overall performance on \textsc{Minimum Wage} marginally. 

\citet{halterman2025codebookllmsevaluatingllms} recommend explicit structural markers, e.g. prefixing each field with ``definition:'' or ``example:''.
In our experiments, when using their structured codebook style (\textit{ii. Structured}, Table~\ref{tab:label_attendance}), \textsc{CCC} shows substantial gains across nearly all metrics, while \textsc{Minimum Wage} and \textsc{Value Creation Science} exhibit significant decreases in all metrics.
We attribute this outcome to the average length of the texts in \textsc{CCC}, which is by far the longest of the three.
Information signaled by markers ("definition") may help the model retrieve relevant information from the full context window.

Following \citet{dunivin2025scaling}'s clause-order finding, we place majority labels first (\textit{iii. Majority Order}).
As shown in Table~\ref{tab:label_attendance}, this ordering yields only minor changes for \textsc{Minimum Wage} and \textsc{Value Creation Science}, but substantially degrades performance on \textsc{CCC}.  Strikingly, \textsc{CCC} achieves its best \textit{micro/macro-F1} when the majority label order is inverted (\textit{iii. Inverted Order}), though this harms performance on the other two datasets. 

These contradictory outcomes across three independent design dimensions may help explain why studies, especially those with limited dataset diversity, sometimes arrive at opposing best-practice recommendations.

\paragraph{Stochastic Instability}
As \citet{Ramanathan_2025} note, LLM (GPT-4) performance can vary across repeated runs.
We observe the same effect with open-weight models when setting temperature above zero and not fixing a random seed.\footnote{Note that temperature 0 reduces but does not fully eliminate non-determinism in hosted settings~\citep{atil-etal-2025-non}, so repeated runs remain informative even under greedy decoding.}
Figure~\ref{fig:rerun_label} shows the aggregated mean F1 scores over 15 such runs.
The degree of instability is dataset- and label-dependent (Figure~\ref{fig:rerun_label}). For \textsc{CCC}, scores remain highly stable. \textsc{Value Creation Science} shows moderate run-to-run differences that are reduced by averaging three runs, while \textsc{Minimum Wage} exhibits stronger variation, particularly for the low-support label \textsc{TURNOVER}. 
The pattern is consistent with the label-level disparities documented in Section~\ref{sec:label_level}: categories with small support are most vulnerable to both systematic underperformance and stochastic noise. 
Our observed variance is modest compared to the greater instability reported by \citet{Ramanathan_2025}, likely because open-weight models allow transparent control over sampling parameters. Closed-source APIs, by contrast, may introduce opaque configuration changes between provider updates, compounding run-to-run variability with silent shifts in model behaviour.

\begin{figure*}
  \centering
  \includegraphics[width=0.8\linewidth]{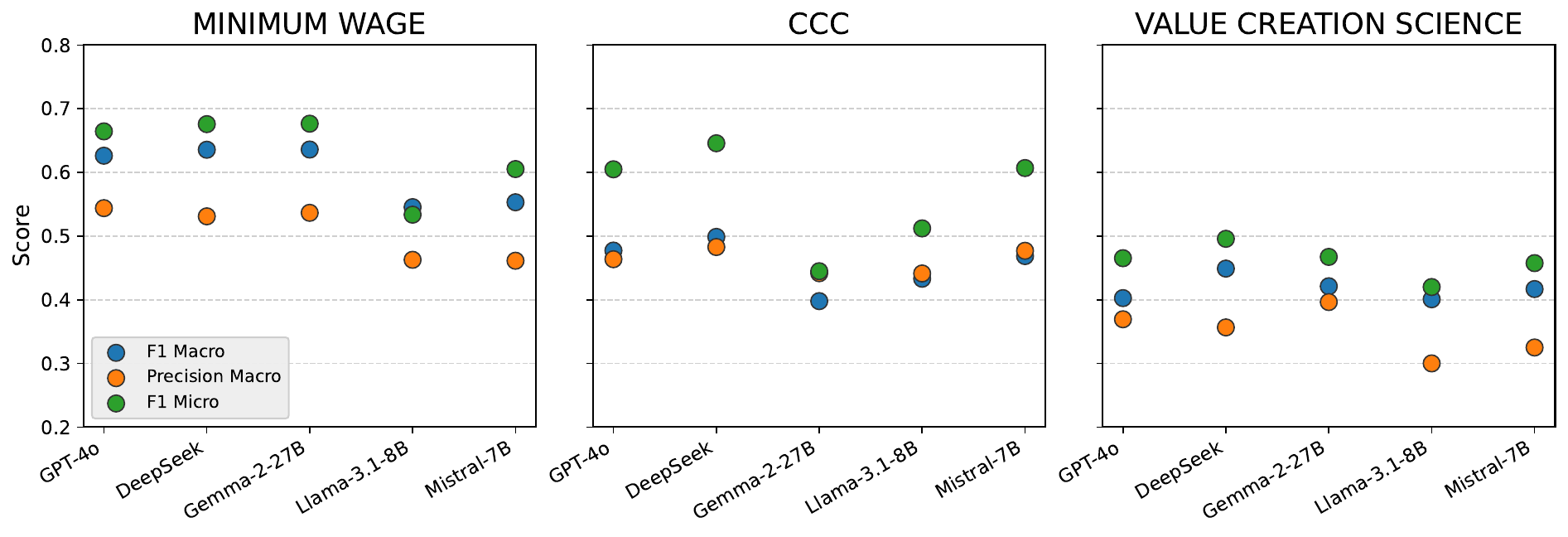}
  \caption{An overview of model performance for the different datasets.}
  \label{fig:model_comp}
\end{figure*}

\begin{table*}
\centering
\resizebox{1.5\columnwidth}{!}{
\begin{tabular}{lccccc}
\hline
 & \textbf{Acc} & \textbf{mi.\ F1} & \textbf{macro-F1} & \textbf{ma.\ P} & \textbf{WLF1} \\
\hline
\textsc{Value Creation Science} 
& $1.9_{1.6}^{2.1}$ 
& $44.9_{44.6}^{45.1}$ 
& $39.6_{39.0}^{40.1}$ 
& $30.5_{29.9}^{31.0}$ 
& $18.8_{15.9}^{21.7}$ \\
\textsc{Value Creation News}        
& $\underline{\mathbf{0.0_{0.0}^{0.1}}}$ 
& $\underline{\mathbf{62.2_{61.7}^{62.8}}}$ 
& $\underline{\mathbf{59.6_{58.9}^{60.2}}}$ 
& $\underline{\mathbf{54.1_{53.5}^{54.7}}}$ 
& $\underline{\mathbf{32.9_{32.2}^{33.5}}}$ \\[2pt]
\hline
\end{tabular}
}
\caption{Metrics over ten runs for \textsc{Value Creation Science} and \textsc{Value Creation News}, shown with their mean value and the 95\% confidence interval over 10 runs. Bold values indicate statistically significant differences under paired $t$-tests ($p<0.05$). Underlined values indicate significance under paired bootstrap tests.}
\label{tab:domain}
\end{table*}

\paragraph{Model Generalization} 
\label{sec:model_gen}
We compare GPT-4o-mini, DeepSeek-V3 \citep{deepseekai2025deepseekv3technicalreport}, Gemma-2-27B-it \citep{gemmateam2024gemma2improvingopen}, Llama-3.1-8B-Instruct, and Mistral-7B-Instruct-v0.3 \citep{jiang2023mistral7b} using the \textit{Unchanged} codebook on all three datasets (Figure~\ref{fig:model_comp}).
No model dominates consistently.
For the \textsc{Value Creation Science} dataset, performance levels are relatively close: Gemma achieves the highest precision, while DeepSeek leads in \textit{micro/macro-F1}. Llama, in contrast, performs weakest overall. On the \textsc{CCC} dataset, Gemma’s F1 scores drop drastically, up to 25 percentage points below the other models, whereas the compact Mistral performs comparably to the much larger DeepSeek and GPT models. The pattern reverses for the \textsc{Minimum Wage} dataset, where Gemma consistently outperforms all others, while Mistral and Llama lag behind.
The absence of a stable model ranking across datasets poses a systemic threat for CSS. If conclusions depend heavily on which model (or version) is chosen, downstream analyses risk being contingent on arbitrary or even opaque implementation details. Closed-source models amplify this problem, as providers update them without notice, leading to silent shifts in performance and prompt sensitivity. What appears to be a stable, replicable pipeline today may yield different results, undermining the credibility of published findings, and may as a consequence even impact policy making.

\paragraph{Domain Transfer Failure}
\label{sec:domain}

A critical assumption in codebook-based content analysis is that coding frameworks remain valid across new data sources within the same theoretical domain. 
We test this assumption by applying the \textsc{Value Creation Science} codebook, developed on academic publications, to news articles covering the same business and economics topics. Both corpora share the underlying constructs but differ in register: style, vocabulary, and communicative conventions change.
The results expose a paradox (Table~\ref{tab:domain}). On one side, the per-label performance improves substantially in the news domain: \textit{micro-F1} rises from 44.89 to 62.22 and \textit{macro-F1} from 39.57 to 59.55, likely because news articles use simpler language and more direct formulations (qualitative examples in Appendix~\ref{app:quali}).
On the other side, this confidence does not extend to a holistic understanding of the coding task: the model fails to predict the full correct label set for any single instance, resulting in an overall \textit{accuracy} of 0.0. 
Such domain sensitivity poses significant risks for comparative analyses and theory testing in CSS. A codebook that performs well on one text type may yield systematically different results on another, even when both sources address the same substantive questions. These inconsistencies can lead to spurious empirical patterns or failed replications.

\renewcommand{\arraystretch}{1.2}
\begin{table*}
\addtolength{\tabcolsep}{-0.1cm}
\centering
\resizebox{\textwidth}{!}{%
\begin{tabular}{l|ccccc|ccccc|ccccc}
\hline
\multirow{2}{*}{\textbf{Optimization}} 
 & \multicolumn{5}{c}{\textbf{\textsc{Minimum Wage}}} 
 & \multicolumn{5}{c}{\textbf{\textsc{CCC}}} 
 & \multicolumn{5}{c}{\textbf{\textsc{Value Creation Science}}} \\
\cline{2-16}
 & \textbf{Acc} & \textbf{mi.\ F1} & \textbf{ma.\ F1} & \textbf{ma.\ P} & \textbf{WLF1} 
 & \textbf{Acc} & \textbf{mi.\ F1} & \textbf{ma.\ F1} & \textbf{ma.\ P} & \textbf{WLF1} 
 & \textbf{Acc} & \textbf{mi.\ F1} & \textbf{ma.\ F1} & \textbf{ma.\ P} & \textbf{WLF1} \\
 
\hline
Unchanged Codebook     
 & 17.5 & 59.8 & 54.4 & 47.1 & 25.0 
 & 38.3 & 40.9 & 38.1 & 44.4 & 7.1
 & 0.7 & 44.5 & 39.4 & 29.8 & 18.2 \\ 
\hline
Label Names  (§ \ref{sec:increase_bias}, § \ref{sec:dataset_dep_opt} )            
 & 26.8 & 56.0 & 51.0 & 44.8 & 22.2 
 & 45.4 & 47.2 & 40.1 & 46.3 & 7.4
 & 3.4 & 42.4 & 35.3 & 29.3 & 8.5 \\ 
Label Definition (§ \ref{sec:increase_bias}, § \ref{sec:dataset_dep_opt} )      
 & 19.1 & 60.0 & 55.7 & 46.1 & 28.6 
 & 52.9 & 55.0 & 44.4 & 45.0 & 13.0
 & 1.9 & 45.5 & 39.8 & 30.8 & 14.8 \\ 
Pos Clarification  (§ \ref{sec:increase_bias}, § \ref{sec:dataset_dep_opt} )      
 & 21.1 & 58.1 & 53.7 & 46.6 & 25.0
 & 35.4 & 37.3 & 35.7 & 44.8 & 7.5
 & 2.6 & 44.5 & 39.0 & 29.4 & 12.3 \\ 
Pos + Neg Clarification (§ \ref{sec:increase_bias}, § \ref{sec:dataset_dep_opt} )
 & 24.7 & 61.1 & 57.5 & 53.9 & 40.0
 & 45.5 & 47.2 & 40.1 & 46.3 & 7.4
 & 2.2 & 44.8 & 40.4 & 30.6 & 23.5 \\ 
Pos Examples (§ \ref{sec:increase_bias}, § \ref{sec:dataset_dep_opt} )            
 & 13.4 & 58.5 & 55.8 & 47.9 & 38.7 
 & 31.3 & 31.6 & 32.4 & 52.1 & 5.1
 & 2.6 & 45.8 & 40.0 & 31.3 & 15.1 \\ 
Pos + Neg Examples (§ \ref{sec:increase_bias}, § \ref{sec:dataset_dep_opt} )  
 & 15.0 & 59.3 & 55.4 & 49.9 & 25.0 
 & 44.2 & 44.9 & 39.4 & 50.7 & 5.7
 & 3.0 & 45.5 & 38.9 & 30.2 & 11.3 \\ 
\hline
Full  (§ \ref{sec:dataset_dep_opt}, § \ref{sec:concept_validity} )            
 & 18.6 & 58.9 & 52.7 & 44.5 & 33.3 
 & 30.9 & 32.2 & 33.4 & 47.3 & 6.4  
 & 3.4 & 43.3 & 38.5 & 28.8 & 13.2  \\ 
APO    (§ \ref{sec:dataset_dep_opt}, § \ref{sec:concept_validity} )     
 & 16.5 & 38.8 & 33.8 & 26.9 & 22.2
 & 57.0 & 60.1 & 45.2 & 45.4 & 9.5  
 & 0.0 & 40.2 & 34.3 & 22.5 & 0.0 \\ 
SCALE  (§ \ref{sec:dataset_dep_opt}, § \ref{sec:concept_validity} )      
 & 17.0 & 39.7 & 33.6 & 22.9 & 9.3
 & 49.5 & 59.9 & 44.0 & 42.2 & 5.6  
 & 0.0 & 38.1 & 36.7 & 23.7 & 26.2 \\ 

\hline
\end{tabular}%
}
\caption{Results for different codebook adaptations across datasets. Each dataset reports accuracy (Acc), micro F1 (mi.\ F1), macro F1 (ma.\ F1), marco precision (ma.\ P), and worst label F1 (WLF1).  For the structured experiment  Mistral-7B-Instruct-v0.3 was used. GPT-4 was used for FULL, APO and SCALE experiments. }
\label{tab:results}
\end{table*}

\section{Failure Modes in Optimization}
\label{sec:optimization}

Failure modes in LLM-based coding do not arise only during inference, additional risks emerge when LLMs are used to optimize codebooks that are later applied during inference. 
To study these effects, we first implement targeted minimal optimizations, introducing small structured changes to the original codebook to test their impact across datasets (see Table~\ref{tab:results}). These perturbations are generated zero-shot in a single LLM rollout. 

Second, we evaluate real-world approaches, in which the complete codebook is optimized.
Specifically, we apply (i) a zero-shot full-codebook improvement (\textit{FULL}), prompting an LLM to refine the entire codebook; (ii) \textit{APO} by \citet{abraham2025promptselectionmattersenhancing}; and (iii) the multi-agent framework \textit{SCALE} \citep{zhao-etal-2025-scale}, both iterative prompt refinement methods.
Using these setups, we analyse potential risks such as concept drift, bias amplification, and the practical feasibility of LLM-based codebook optimization.

\subsection{Amplified Bias and Sensitivity}
\paragraph{Increasing Label Bias}
\label{sec:increase_bias}

The label-level disparities observed during inference (Section~\ref{sec:label_level}) are amplified when LLMs are used to optimize codebooks. Small modifications to definitions or examples can disproportionately shift performance across labels (see Table~\ref{tab:results}). 
When evaluation relies solely on aggregate metrics, such as \textit{micro-F1}, there is a danger of endorsing adaptations that improve majority classes while harming minority ones.
A concrete case illustrates the danger: adding positive and negative examples improves \textit{micro/macro-F1}, \textit{accuracy}, and \textit{macro-Precision} on both \textsc{CCC} and \textsc{Value Creation Science}, yet simultaneously drives \textit{worst-label-F1} down to 5.67 and 11.32 respectively. Judged by aggregate metrics alone, these optimizations would be endorsed as beneficial. In practice, they systematically suppress minority categories, precisely the labels most vulnerable to begin with. The problem is compounded by automatic methods such as APO, which optimize for \textit{micro-F1} by design and therefore have no mechanism to prevent minority-label degradation.

\begin{table*}
\centering
\small
\renewcommand{\arraystretch}{1.1}
\setlength{\tabcolsep}{8pt}
\begin{tabular}{@{}lp{14cm}@{}}
\toprule
\textbf{Method} & \textbf{Definition for the \textsc{REVENUE} label} \\ 
\midrule
\textsc{Original} &
The benefit or reward an actor receives for contributing to value creation. Can be tangible or intangible. \textit{Examples:} wages, profits, royalties, increases in social capital.
\\
\hline
\textsc{Full} &
The reward or advantage received by an actor for participating in value creation, either tangible or intangible. \textit{Examples:} wages, profits, royalties, increases in social capital.\\[0.6em]

\textsc{Apo } &
The benefit or compensation received by an actor for their involvement in value creation. This can be either tangible or intangible. \textit{Examples:} wages, profits, royalties, increases in social capital.\\[0.6em]

\textsc{Scale} &
The benefit or reward an actor receives for contributing to value creation. Can be tangible or intangible. \textit{Examples:} wages, profits, royalties, increases in social capital, financial implications of operational changes, and benefits from educational initiatives. \textit{Clarification:} Revenue can be seen as the return on investment for actors involved in the value creation process. \\

\bottomrule
\end{tabular}
\caption{Optimized versions of the label \textsc{REVENUE} across three approaches.}
\label{tab:actor_optimization}
\end{table*}

\paragraph{Dataset Dependency for Optimization}
\label{sec:dataset_dep_opt}
The dataset sensitivity documented for inference (Section~\ref{sec:guidelines}) intensifies when LLMs are used to optimize codebooks, since each refinement step compounds the interaction between prompt wording and data characteristics. The absence of universal optimization rules motivates automated methods such as APO and SCALE, yet these introduce systematic risks of their own.

Our experiments show that even minimal adjustments, like adding examples, inserting clarifications, or reformulating definitions produce inconsistent effects across datasets (Table \ref{tab:results}). What improves performance in one setting often degrades it in another, casting doubt on the existence of generalizable strategies for LLM-based refinements. This underscores the need for dataset-specific optimization methods that adapt prompts automatically rather than relying on fixed heuristics.
Among real-world optimization methods, the simple FULL approach proves most stable, with \textit{macro-F1} decreasing by at most 5 points across datasets (Table \ref{tab:results}).
 APO is more volatile: it performs well on \textsc{CCC}, but for \textsc{Value Creation Science} it removed one label entirely, leading to weaker results. 
 SCALE also exhibits high variability, improving \textsc{CCC} performance but reducing most metrics for the other two datasets. For \textsc{Value Creation Science}, repeated SCALE runs produced drastically different and inconsistent outcomes ranging from leaving all labels unchanged to completely rewriting the codebook, resulting in labels that bore little resemblance to the originals.
 This run-level instability means that even the choice of which optimization run to deploy becomes a consequential, yet essentially arbitrary, decision.

\begin{figure}
  \centering
  \includegraphics[width=1.0\linewidth]{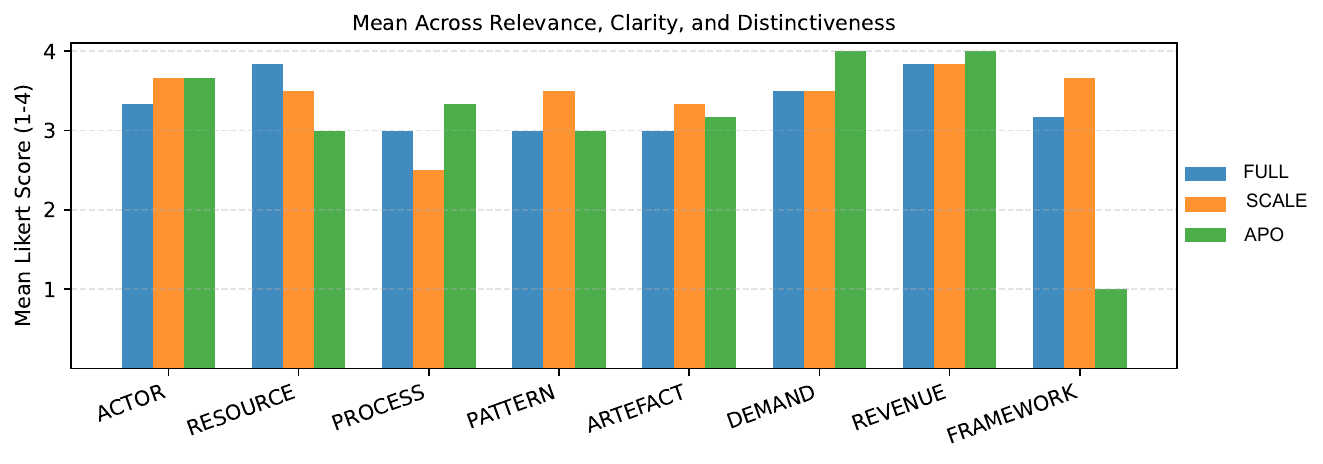}
  \caption{Overview of mean Likert scores across dimensions for each optimization method. }
  \label{fig:apo_scale}
\end{figure}

\begin{table}
\centering
\small
\setlength{\tabcolsep}{1.5pt}      
\renewcommand{\arraystretch}{0.55} 
\setlength{\aboverulesep}{1pt}   
\setlength{\belowrulesep}{1pt}   
\setlength{\belowbottomsep}{1pt} 

\begin{tabular}{lccc}
\toprule
\textbf{Dataset} & \textbf{Full} & \textbf{APO} & \textbf{SCALE} \\
\midrule
\textsc{Minimum Wage} (§ \ref{sec:efficiency})  & 1 & 113,200 & 856 \\
\textsc{CCC} (§ \ref{sec:efficiency}) & 1 & 113,200 & 688 \\
\textsc{Value Creation Science} (§ \ref{sec:efficiency}) & 1 & 113,200 & 1,044 \\
\bottomrule
\end{tabular}
\caption{Number of API calls required for each optimization method across datasets (§\ref{sec:efficiency}).}
\label{tab:api_calls}
\end{table}

\subsection{Conceptual Validity of Optimized Codebooks}
\label{sec:concept_validity}

Optimization alters label definitions and occasionally label names, raising the question of whether the resulting codebooks still capture the theoretical constructs encoded in the original design. 
To evaluate this, two domain experts assessed each label description in the \textsc{Value Creation Science} dataset using a 1–4 Likert scale \citep{Lynn1986DeterminationAQ}, a method commonly used to evaluate natural language generation outputs \citep{amidei-etal-2019-use}. Due to low rating variance, inter-annotator agreement measured by Cohen’s $\kappa$ is relatively low ($\kappa = 0.34$), although the observed agreement between annotators is high: 51.6\% exact agreement and 87.3\% of ratings differing by at most one scale point.\footnote{Cohen’s $\kappa$ adjusts observed agreement by the agreement expected by chance. When rating variance is low (i.e., both annotators tend to use the same rating categories very frequently), the expected chance agreement becomes high. This reduces the resulting $\kappa$ value even if the annotators largely agree in practice. In our case, both annotators consistently assigned high scores, and one rater frequently used the maximum score (4), which increases expected chance agreement and lowers $\kappa$.} Three dimensions were evaluated: \textit{relevance}, \textit{clarity}, \textit{distinctiveness} (see Appendix~\ref{app:critera}).

\paragraph{Relevance}
For relevance, the expert evaluated whether each label still reflected the intended concept of the original codebook. The highest mean score was achieved by SCALE (3.56), followed by FULL (3.44) and APO (3.19)  (Figure~\ref{fig:apo_scale_small}). The APO approach received consistently high ratings, except for the label \textit{FRAMEWORK}, which was removed during optimization. Despite this, overall performance in Table~\ref{tab:results} remains strong given that one label could not be detected. %

\paragraph{Clarity}
Since codebooks are intended for downstream analysis, it is essential that optimized label descriptions remain interpretable and clear to social scientists.
Both SCALE outperforms FULL and APO with mean clarity scores of 3.44 compared to 3.25 and 3.06. This gap may stem from the richer definitions and examples that help differentiate labels (see Table~\ref{tab:actor_optimization}).

\paragraph{Distinctiveness}
For codebooks to be analytically useful, labels must remain clearly distinct from one another. 
Again, FULL and SCALE yield higher mean Likert scores than APO. 

\paragraph{}
For \textit{clarity} and \textit{relevance}, SCALE produces definitions experts judge superior to FULL (detailed scores in Appendix~\ref{app:likert}. Yet FULL often achieves higher F1 scores (Table~\ref{tab:results}), revealing a disconnect: predictive gains can arise from simplified definitions that better align with model behavior while reducing conceptual interpretability. This tension highlights that quantitative metrics alone are insufficient for CSS applications where codebooks encode theoretical constructs.

\subsection{Efficiency and  Accessibility}
\label{sec:efficiency}

Beyond performance, optimization approaches must also be feasible for researchers in practice. We therefore compare the efficiency of different methods by counting the number of API calls required for each approach (see Table~\ref{tab:api_calls}). This provides an approximate measure of computational cost.
FULL requires a single call per dataset. SCALE uses 688--1,044 calls depending on the dataset. APO requires 113,200 calls uniformly, over two orders of magnitude more than SCALE and five orders more than FULL, yet does not consistently outperform either alternative (Table~\ref{tab:results}). The cost-benefit profile of iterative optimization is thus far from guaranteed. Beyond cost, accessibility also matters: minimal prompt adjustments are within reach of any social scientist, whereas APO and SCALE demand non-trivial engineering skills.

\section{Minimal Validation Protocol}
\label{sec:protocol}
To mitigate failure modes from Sections~\ref{sec:inference} and~\ref{sec:optimization}, we recommend the following minimal, low-cost validation protocol, intended as a practical screening procedure rather than a statistically exhaustive evaluation:
\begin{tcolorbox}[colback=gray!10,colframe=black,title=\textbf{Minimal Validation Protocol}, 
left=2pt,right=3pt,top=1pt,bottom=1pt,boxsep=2pt]
\setlist[itemize]{itemsep=1pt, topsep=2pt,leftmargin=1em}

\textbf{Tail Risk:} Report \textit{macro-F1}, \textit{worst-label-F1}, per label scores and label supports alongside micro-F1/accuracy; include \textit{macro-Precision} to detect overcoding.  

\textbf{Meaning Adherence:} Test \textit{Label-Only}, \textit{Generic Label}, and \textit{Swapped Label} to detect reliance on label-name semantics.  

\textbf{Stability:} Repeat inference $\geq 3$  as a practical minimum for detecting substantial instability; report the number of runs and the resulting variability, and use more runs  when variance is high or the API is stochastic/opaque. 

\textbf{Sensitivity:} Apply  semantics-preserving formatting or label-order perturbations to reveal prompt-specific effects.  

\textbf{Transfer:} If possible, evaluate on an additional dataset from the same domain to detect shifts under domain or style changes.  

\textbf{Optimization Safeguards:} When refining codebooks, include worst-label-F1 to protect minority labels, verify constructs with experts, and report computational cost.
\end{tcolorbox}

\section{Conclusion}
LLMs enable large-scale content analysis in CSS but remain fragile and context-dependent.  Our experiments demonstrate that high aggregate scores routinely coexist with severe label-level disparities, reliance on label-name semantics over codebook definitions, instability across runs, sensitivity to superficial formatting choices, and fragile domain transfer. When LLMs are additionally used to optimize codebooks, these problems do not diminish, they compound: optimization amplifies minority-label bias, produces dataset-contingent effects, and can silently alter the theoretical constructs a codebook is designed to measure.
A recurring theme across all failure modes is that standard metrics are insufficient to detect them. To address this, we proposed a minimal validation protocol (Section~\ref{sec:protocol}) that operationalizes our findings into concrete checks, intended to be lightweight enough for routine adoption in CSS research.

\section*{Limitations}

Our analysis is limited to a small set of datasets, codebooks, and model configurations. While these reflect common CSS research settings, results may not generalize across domains, languages, or annotation frameworks. Nonetheless, our evaluation framework is applicable to any dataset, codebook, or LLM, and  we hope future work will extend it to larger, multilingual, and domain-specific contexts. Finally, the human evaluation of optimized codebooks relies on only two experts, which may limit intersubjective validity.

\section*{Ethical Considerations}

This work studies the reliability of LLM-based text annotation and highlights potential risks when such annotations are used in downstream social science analyses. By identifying failure modes in inference and codebook optimization, our goal is to promote more transparent and responsible use of LLMs as measurement tools in research workflows.
All datasets used in this study are either publicly available or released by the authors with appropriate documentation. We do not introduce new personal or sensitive data, and our experiments focus on methodological evaluation rather than producing judgments about individuals or groups.
LLMs were used to assist with grammar, spelling, and refinement of the manuscript text. All technical content, analyses, and conclusions were developed and verified by the authors.

\section*{Acknowledgments}
We thank the reviewers for their insightful feedback. We thank Jan-Peter Bergmann for computing infrastructure, support and advice. ES and YC acknowledges financial support funded by the German Federal Ministry of Research, Technology and Space (BMFTR) under grant Wertschöpfungsradar 02J21A00X. We thank all project partners for their collaboration and the Scientific Advisory Board for their valuable guidance and support. BP acknowledges support funded by ERC Consolidator Grant DIALECT 101043235.


\bibliography{custom}

\appendix

\section{Codebooks}
\label{app:codebooks}

We retained the original codebooks as closely as possible and reformatted them into a standardized prompt format for annotation.

\subsection*{\textsc{Value Creation} Codebook Prompt}

\begin{quote}
\small
The value creation codebook prompt:

You are an economic analyst examining the socio-technical structure of value creation systems. You have been handed the following codebook as instructions to code text units for their aspects of value creation.

\medskip

\textbf{ Aspect Annotation in short text units.}

The task is to find the aspect(s) present in a text unit. An aspect is defined as an element of the value creation system being discussed. It is possible that more than one aspect applies to a text unit; please mark all applicable aspects.

Below, the aspect categories are further explained. Each category has a label name and details defining or illustrating them.

\medskip

\textbf{[ACTOR]}
\begin{itemize}
  \item A person or group providing a resource to help create value. Can be individuals or organized/informal groups (e.g., communities, companies).
  \item Examples: manufacturers, investors, employees, innovators, user communities.
\end{itemize}

\textbf{[RESOURCE]}
\begin{itemize}
  \item Something an actor uses to create value. Can be materials, objects processed, tools, information, or skills.
  \item Examples: machinery, raw materials, financial capital, data, technology, expertise.
\end{itemize}

\textbf{[PROCESS]}
\begin{itemize}
  \item A series of actions where actors use resources to create value. Includes primary activities (making and delivering products) and supporting activities.
  \item Examples: manufacturing, procurement, distribution, maintenance, servicing.
\end{itemize}

\textbf{[PATTERN]}
\begin{itemize}
  \item A regular and repeated way value creation is performed. Encompasses rules, norms, routines, and practices shaping actor roles or constraining actions.
  \item Examples: organizational structures, communication protocols, property rights, quality standards.
\end{itemize}

\textbf{[ARTEFACT]}
\begin{itemize}
  \item The output of the value creation process. Can be physical goods, intangible services, or product-service systems.
  \item Examples: consumer products, digital services, experiences, entertainment offerings.
\end{itemize}

\textbf{[DEMAND]}
\begin{itemize}
  \item A stated or expected need that an artefact fulfills. Needs can originate from individuals, groups, or society as a whole.
  \item Examples: consumer demand for essentials, societal challenges, personal preferences.
\end{itemize}

\textbf{[REVENUE]}
\begin{itemize}
  \item The benefit or reward an actor receives for contributing to value creation. Can be tangible or intangible.
  \item Examples: wages, profits, royalties, increases in social capital.
\end{itemize}

\textbf{[FRAMEWORK]}
\begin{itemize}
  \item External factors influencing how value is created. Includes environmental, political, legal, and social factors.
  \item Examples: regulatory regimes, market conditions, intellectual property laws, educational infrastructure.
\end{itemize}

In some cases, more than one label applies.  
Do not mention labels that do not apply.

\medskip

\textbf{\# Task}

Assign one or more of the labels from the codebook to the following text:
\end{quote}

\subsection*{\textsc{Minimum Wage} Codebook Prompt}

\begin{quote}
\small
You are a well-trained social scientist with a speciality for coding data for scientific research.
You have been handed the following markdown codebook as instructions to code example sentences for their aspects.

\medskip

\textbf{ Aspect Annotation in short argumentative text units.}

The task is to find the aspect of an argumentative sentence or short text unit.
An aspect in this task is defined as a sub-topic of discourse in the broader
topic of the debate about minimum wage.
It is possible,
that more than one aspect is present in a sentence, please mark all applicable
aspects.

Below, the aspect categories are further explained. Each category has a label name and details defining or illustrating them.

\medskip

\textbf{[MOTIVATION/CHANCES]}
\begin{itemize}
  \item Incentives and opportunities for workers to start a job/career, futher educate themselves and move up in a company.
  \item Influence of minimum wage on education in general or for particular groups.
  \item Eagerness and aspiration of workers.
  \item Keywords: motivation, incentives to education, upward mobility, worker morale.
  \item Examples:
  \begin{itemize}
    \item "If we raise the minimum wage, low skilled workers might become complacent and do not want to move up"
    \item "Paying employee better will make them more motivated to work for your company."
  \end{itemize}
\end{itemize}

\textbf{[COMPETITION/BUSINESS CHALLENGES]}
\begin{itemize}
  \item Effects of minimum wage on business competition and profitability.
  \item Challenges for small businesses.
  \item Keywords: labour costs, payability, SME competition.
\end{itemize}

\textbf{[PRICES]}
\begin{itemize}
  \item Effects on consumer prices and inflation.
  \item Keywords: inflation, living costs, housing, daycare.
  \item Example: "Higher prices cancel positive effects of minimum wage."
\end{itemize}

\textbf{[SOCIAL JUSTICE/INJUSTICE]}
\begin{itemize}
  \item Fairness, inequality, poverty, and living wage.
  \item Keywords: dignity, standard of living, poverty line.
  \item Examples:
  \begin{itemize}
    \item "Workers should be able to afford paying their bills"
    \item "Wages have to fairly compensate workers"
  \end{itemize}
\end{itemize}

\textbf{[WELFARE]}
\begin{itemize}
  \item Relationship between minimum wage and welfare spending.
  \item Keywords: welfare, subsidies, tax credits.
  \item Example: "Full-time workers should not have to rely on welfare"
\end{itemize}

\textbf{[ECONOMIC IMPACT]}
\begin{itemize}
  \item Effects on the overall economy and consumer spending.
  \item Keywords: ripple effect, economic growth.
\end{itemize}

\textbf{[TURNOVER]}
\begin{itemize}
  \item Effects on employee turnover and training costs.
  \item Keywords: job stability, hiring costs.
\end{itemize}

\textbf{[CAPITAL VS LABOR]}
\begin{itemize}
  \item Power disparity between workers and owners.
  \item Role of unions and exploitation.
  \item Keywords: collective bargaining, working class, sweatshops.
\end{itemize}

\textbf{[GOVERNMENT]}
\begin{itemize}
  \item State intervention and wage legislation.
  \item Keywords: regulation, free market.
  \item Example: "State vs. free market."
\end{itemize}

\textbf{[UN/EMPLOYMENT RATE]}
\begin{itemize}
  \item Effects on employment, unemployment, automation, outsourcing.
  \item Keywords: job loss, labor supply.
\end{itemize}

\textbf{[LOW-SKILLED]}
\begin{itemize}
  \item Effects on low-skilled and entry-level workers.
  \item Distinction from youth workers.
  \item Keywords: unskilled, entry-level.
\end{itemize}

\textbf{[YOUTH AND SECONDARY WAGE EARNERS]}
\begin{itemize}
  \item Part-time workers, students, teenagers.
  \item Secondary income earners.
\end{itemize}

In some cases, more than one label applies.  
Do not mention labels that do not apply.

\medskip

\textbf{\# Task}

Assign one or more of the labels from the codebook to the following text:
\end{quote}

\subsection*{\textsc{CCC} Protest Dataset Prompt}

\begin{quote}
\small
Instruction: You're an expert political scientist categorizing news stories about protests into categories. Carefully read the definitions below, read the story, and write the single Label that best match the story.

\medskip

\textbf{[PROTEST]}
\begin{itemize}
  \item A crowd gathering to express disagreement with, or disapproval or anger or frustration toward, a specific individual or organization that is at or near the crowd’s gathering point.
  \item The crowd must be near the target of the protest or in negative reaction to a recent or current event.
\end{itemize}

\textbf{[RALLY]}
\begin{itemize}
  \item A crowd gathering to demonstrate and reinforce their mutual support for a set of political demands or claims.
  \item Typically includes speeches and may support press conferences.
\end{itemize}

\textbf{[DEMONSTRATION]}
\begin{itemize}
  \item A crowd gathering to demonstrate their support to an external audience.
  \item Usually involves signs, banners, flags, or pickets.
\end{itemize}

\textbf{[MARCH]}
\begin{itemize}
  \item A crowd moving on foot from one location to another.
  \item Does not include motorized or bicycle movement.
\end{itemize}

\medskip

\textbf{\# Task (single-label multi-class)}

Do not write anything else except the Label in all caps.

Output Reminder: Write the name of the Label that fits best, with no other text. For example, "Label: PROTEST", "Label: RALLY", etc.

\medskip

\textbf{\#\# Text}
\end{quote}

\section{Details on Dataset Creation}
\label{app:datasets}

This appendix describes the construction of the datasets\textsc{Value Creation Science} and \textsc{Value Creation News}. The datasets were developed as part of the \textit{Wertsch\"opfungsradar} (Value Creation Radar, WSR) research project.

\subsection{Original Dataset}
\subsubsection{Annotation Scheme}
\label{app:scheme}

The annotation scheme is grounded in an abstract model of value creation systems, which distinguishes eight interdependent categories that characterise the essential elements of value creation processes. Each category is defined relationally---that is, through its connections to other elements in the model. The categories and their definitions are as follows:

\begin{itemize}[itemsep=0.5pt, topsep=1pt]
    \item \textbf{Actor}~(\textsf{0}): an individual or a collective that contributes a resource at its disposal to a process of creating an artefact.
    \item \textbf{Resource}~(\textsf{1}): a capability of an actor that is used in the process of creating an artefact.
    \item \textbf{Process}~(\textsf{2}): an activity for the creation of an artefact to which actors contribute their resources.
    \item \textbf{Pattern}~(\textsf{3}): an organising principle through which a process integrates the resources to create an artefact.
    \item \textbf{Artefact}~(\textsf{4}): a good that is created by integrating resources according to a pattern and that serves a demand.
    \item \textbf{Demand}~(\textsf{5}): an articulated or anticipated need that an artefact serves.
    \item \textbf{Revenue}~(\textsf{6}): a return an actor receives for contributing to the process of creating an artefact.
    \item \textbf{Framework}~(\textsf{7}): a context factor that influences how value creation takes place.
\end{itemize}

Annotators were instructed to label noun phrases in English-language abstracts of scientific articles about value creation. Labels capture both positive and negative instances of each category (e.g., financial losses are labelled \textsc{Revenue}; material costs are labelled \textsc{Resource}).

\subsubsection{Annotation Process}
\label{app:process}

The annotation procedure was developed iteratively across two phases: a pilot study and a main annotation campaign.

\paragraph{Pilot study.} An internal pilot was conducted in June--July~2023, in which 8~annotators labelled 20~abstracts in parallel using the \textsc{Doccano} annotation tool.\footnote{\url{https://github.com/doccano/doccano}} The pilot yielded a Fleiss'~$\kappa$ of 0.31, indicating fair for multi-label span annotation, but insufficient agreement. A qualitative analysis of 75~annotator comments and a co-occurrence analysis of category overlaps revealed three main sources of disagreement: (i)~ambiguous category boundaries, particularly between \textsc{Pattern} and \textsc{Process}, \textsc{Revenue} and \textsc{Demand}, and \textsc{Actor} and \textsc{Framework}; (ii)~missing formal coding rules regarding grammatical forms, compound phrases, and abbreviations; and (iii)~insufficient attention to the value-creation context when annotating.

\paragraph{Guideline revision.} Based on the pilot findings, the annotation guidelines were substantially revised. Key improvements included: sharpened category definitions supplemented with illustrative positive and negative examples drawn from the pilot data; the introduction of a step-by-step annotation procedure recommending that annotators first identify \textsc{Actor}, \textsc{Process}, and \textsc{Artefact} before turning to the remaining categories; explicit coding rules restricting annotations to complete noun phrases, prohibiting the splitting of compound expressions, and requiring annotators to consider the relationships between labelled entities as specified by the value creation model. A one-page reference sheet summarising the definitions, explanatory notes, and examples was provided as a quick-reference document.

\paragraph{Annotator training.} Before the main campaign, annotators participated in a training session consisting of a presentation of the revised guidelines followed by a joint practice labelling exercise. An interactive training quiz was also distributed, which walked annotators through a sample text and tested their understanding of noun phrase boundaries, context-dependent category assignment, and the distinction between entities and concepts.

\paragraph{Main annotation campaign.} The main campaign began in September~2023. Annotators labelled abstracts using \textsc{Doccano} with the eight predefined categories. The final dataset comprises 921~annotated abstracts.
The collected annotated spans were proofed by a domain expert in order to improve label quality and remove any ambiguous data points.All involved social scientists consented to the use and release of this dataset.





\subsection{\textsc{Value Creation Science}}
From these 921 annotated abstracts, we selected the sentences which contained annotated noun-phrases and transformed the data into a multi-label dataset.
That is, the whole sentence gets the labels of the identified spans that are contained in it.
After that, we selected only sentences with at least one label and which carried enough context for the labels they carried.
This helps to minimize context loss in the dataset transformation.

\subsection{\textsc{Value Creation News}}
For the VALUE CREATION NEWS dataset, we collected 60 sentences from business-related news feeds and had the three most reliable annotators from the pilot study to label those.







\section{Qualitative Examples of Domain Differences}
\label{app:quali}

Below we provide example sentences from the \textsc{Value Creation News} dataset. Compared to scientific abstracts in the \textsc{Value Creation Science} dataset, these texts typically use simpler language and more direct formulations.

\begin{enumerate}[itemsep=0.5pt, topsep=1pt]
    \item \textbf{Example 1:} “GameStop has significantly reduced its investments in cryptocurrencies, NFTs, and Web3 applications, now holding no crypto tokens and shifting focus back to core business areas.”
    
    \item \textbf{Example 2:} “The telegraph quickly became popular for casual communication, and Ezra Cornell led its construction through southern New York.”
    
    \item \textbf{Example 3:} “Modern companies must adopt data-driven strategies, leveraging technologies like IoT, Big Data, and AI to remain competitive and optimize decision-making.”
\end{enumerate}


\section{Statistical Significance Testing}
\label{app:significance_checklist}

In the following, we present the procedure for assessing (i)~run-to-run stability and (ii)~instance-level uncertainty when comparing each prompt prototype against the baseline.

\subsection{Setup}

For each dataset, we compare each prompt prototype against a fixed baseline prompt. We perform $R{=}10$ independent inference runs per (dataset, system) pair and store all test-set predictions.
We compute accuracy, macro-F1, micro-F1, macro-precision, and micro-precision per run. Multi-label accuracy is subset accuracy (exact match). We set \texttt{zero\_division=0} for F1 and precision.

\subsection{Run-Level Stability}

We assess robustness to inference stochasticity via paired comparisons within each run index. For each run $r \in \{1,\dots,R\}$, we compute metric values $M^{(r)}_{\text{base}}$ and $M^{(r)}_{\text{proto}}$ on the full test set, then form deltas $\delta_r = M^{(r)}_{\text{proto}} - M^{(r)}_{\text{base}}$. We apply a two-sided one-sample $t$-test on $\{\delta_r\}_{r=1}^{R}$ to test $H_0{:}\;\mathbb{E}[\delta]=0$, and report per-system mean$\pm$std, $\bar{\delta}$, the $t$-statistic, $p$-value, and $t$-based 95\% confidence intervals with $R{-}1$ degrees of freedom. As a non-parametric sensitivity check, we also compute a two-sided Wilcoxon signed-rank test.

\subsection{Instance-Level Uncertainty}

We quantify finite-sample uncertainty via a paired bootstrap that preserves baseline--prototype pairing on the same sampled instances ($B{=}20{,}000$; seed${=}42$; 95\% confidence level).

The observed run-averaged improvement is $\Delta_{\text{obs}} = \frac{1}{R}\sum_{r} (M^{(r)}_{\text{proto}} - M^{(r)}_{\text{base}})$. For each replicate~$b$, we resample $N$ instance indices with replacement and apply this index vector to $y$, $\hat{y}^{(r)}_{\text{base}}$, and $\hat{y}^{(r)}_{\text{proto}}$ for every run~$r$, yielding a run-averaged bootstrap delta $\Delta_b = \frac{1}{R}\sum_{r} (M^{(r)}_{\text{proto},b} - M^{(r)}_{\text{base},b})$. We report the 95\% percentile CI $[\Delta_{2.5}, \Delta_{97.5}]$ and a two-sided bootstrap $p$-value:
\[
  p_{\text{boot}} = 2\min\!\bigl(\Pr(\Delta_b \le 0),\; \Pr(\Delta_b \ge 0)\bigr).
\]



\subsection{Out-of-Domain Comparison}

For the dataset-shift analysis, the two test sets contain different instances and are not paired at the instance level. We compute per-run metrics on each test set and apply a paired-delta $t$-test across runs. We obtain a 95\% CI and two-sided $p$-value via an unpaired bootstrap over runs (same $B$ and seed as above). For worst-label analysis, we compute per-label F1 per run, identify the label with the lowest mean F1 per test set, and compare these distributions using Welch's $t$-test and an unpaired run-level bootstrap.

\section{Encoder Baseline: Fine-Tuned RoBERTa}
\label{app:roberta_baseline}

We additionally evaluate a fine-tuned RoBERTa \cite{liu2019roberta} baseline to contextualize the observed failure modes against a conventional supervised encoder model.
We fine-tune \texttt{roberta-large}\footnote{https://huggingface.co/FacebookAI/roberta-large} as a multi-label, or multi-class for CCC, sequence classification model using the training split of each dataset and select the best checkpoint on the validation split. Inputs are truncated to a maximum length of 512 tokens. We train for 3 epochs with a learning rate of $2 \times 10^{-5}$, batch size 8.

Table~\ref{tab:roberta_baseline} reports the performance of the RoBERTa baseline across the three datasets. The results show that RoBERTa exhibits a pattern similar to the one observed for LLMs: relatively high aggregate scores, especially Micro-F1, can coexist with substantially lower Macro-F1 and a Worst-Label-F1 of 0.0. What shows that even a supervised encoder baseline can achieve moderate Micro-F1 while entirely failing on minority or difficult labels. However, the underlying causes of these failures differ. In LLM-based coding pipelines, failures may arise from prompt sensitivity, run instability, domain sensitivity, or semantic drift in label interpretation. In contrast, RoBERTa's collapse on individual labels is more likely driven by label imbalance and sparse supervision, which are common in content analysis datasets.

\begin{table}
\centering
\resizebox{\columnwidth}{!}{

\begin{tabular}{lccccc}
\toprule
\textbf{Dataset} & \textbf{Acc.} & \textbf{Micro-F1} & \textbf{Macro-F1} & \textbf{Macro-Prec.} & \textbf{WLF1} \\
\midrule
CCC                    & 72.67 & 73.95 & 49.51 & 51.32 & 0.00 \\
Minimum Wage           & 52.58 & 67.67 & 36.65 & 53.29 & 0.00 \\
Value Creation Science & 11.15 & 31.38 & 13.46 & 18.21 & 0.00 \\
\bottomrule
\end{tabular}
}
\caption{Performance of the fine-tuned RoBERTa-large baseline across datasets.}
\label{tab:roberta_baseline}
\end{table}

\section{Detailed Likert Scores}
\label{app:likert}
Figure~\ref{fig:apo_scale} shows the Likert scores per label for each dimension.
\begin{figure*}[t]
  \centering
  \includegraphics[width=1.0\linewidth]{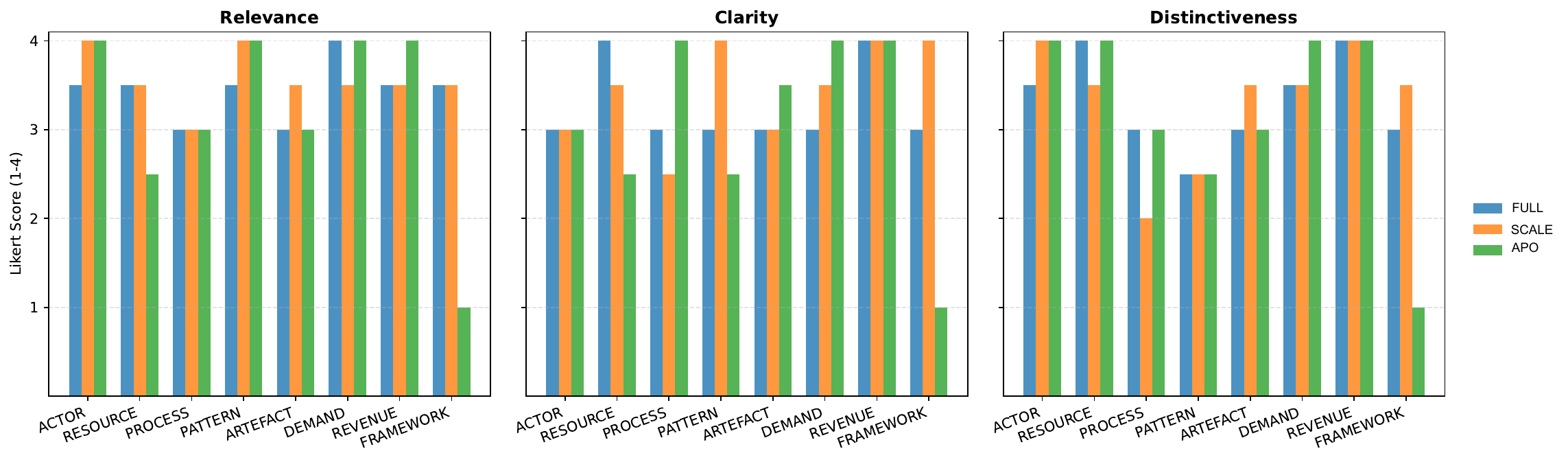}
  \caption{Overview of relevance, clarity, and distinctiveness across optimization methods. }
  \label{fig:apo_scale_small}
\end{figure*}

\section{Expert Evaluation Criteria}
\label{app:critera}

\textbf{Likert Scale}
\begin{itemize}[itemsep=0.5pt, topsep=1pt]
    \item \textbf{1} – Not relevant / clear / distinctive
    \item \textbf{2} – Somewhat relevant / clear / distinctive
    \item \textbf{3} – Quite relevant / clear / distinctive
    \item \textbf{4} – Highly relevant / clear / distinctive
\end{itemize}

\textbf{Evaluation Dimensions}

\begin{itemize}[itemsep=0.5pt, topsep=1pt]
    \item \textbf{Relevance:} Does the code or label still capture the intended concept from the original codebook?
    
    \item \textbf{Clarity:} How easily can the code/label and its definition be understood by a researcher applying it?
    
    \item \textbf{Distinctiveness:} Is the code/label sufficiently distinct from other codes in the set, or could it overlap or be confused with another?
\end{itemize}

\end{document}